\documentclass[letterpaper, 10 pt, conference]{ieeeconf}  % Comment this line out if you need a4paper
\IEEEoverridecommandlockouts
\usepackage{cite}
\usepackage{comment}
\usepackage{balance}
\usepackage[table]{xcolor}
\usepackage{amsmath,amssymb,amsfonts}
\usepackage{subfigure}
\usepackage{graphicx}
\usepackage{comment}
\usepackage{textcomp}
\usepackage{multirow} 
\usepackage{xcolor}
\usepackage{float}
\usepackage{tabularray}
\usepackage{hyperref}
\usepackage{array}
\usepackage{booktabs}
\usepackage{verbatim} 
\usepackage{makecell} % For line breaks in table cells
\usepackage{algorithm}
\usepackage{algpseudocode}
\usepackage{bm} % For bold math symbols

\makeatletter
\usepackage[%
    top=0.78in,       % 上边距
    bottom=0.78in,    % 下边距
    left=0.78in,      % 左边距
    right=0.78in      % 右边距
]{geometry}

\newcommand{\Rmnum}[1]{\expandafter\@slowromancap\romannumeral #1@}

\makeatother
\def\BibTeX{{\rm B\kern-.05em{\sc i\kern-.025em b}\kern-.08em
    T\kern-.1667em\lower.7ex\hbox{E}\kern-.125emX}}

\title{\LARGE \bf Learning Expert Strategy for Autonomous Robotic Endovascular Intervention via Decoupled Procedural Execution}

\author{Yanxi Chen$^{1,2}$, Tianliang Yao$^{3}$, Shaolong Tang$^{1}$, Jiyuan Zhao$^{1}$, Hengyu Hu$^{1,4}$, \\ Zhaoxing Li$^{1}$, Antonio Sánchez Egea$^{5}$, Peng Qi$^{1, *}$
\thanks{This work has been accepted by IEEE/RSJ IROS 2026. Copyright may be transferred without notice, after which this version may no longer be accessible.}
\thanks{This work is supported by the National Key Research and Development Program of China under Grant No. 2023YFB4705200, and the National Natural Science Foundation of China under Grant No. 52575034. The authors would like to thank Mr. Tao Liu from Shanghai Operation Robot Co., Ltd., for providing technical support in experiments. \emph{(*Corresponding Author: Peng Qi, email: pqi@tongji.edu.cn)}.}% <-this % stops a space
\thanks{$^{1}$Department of Control Science and Engineering, College of Electronics and Information Engineering, and Shanghai Institute of Intelligent Science and Technology, Tongji University, Shanghai 200092, China;}
\thanks{$^{2}$School of Mechanical Engineering, Tongji University, Shanghai 200092, China;}
\thanks{$^{3}$Department of Electronic Engineering, Faculty of Engineering, The Chinese University of Hong Kong, Hong Kong SAR 999077, China;}
\thanks{$^{4}$School of Mechatronic Engineering and Automation, Shanghai University, Shanghai 200444, China;}
\thanks{$^{5}$Department of Mechanical Engineering, Universitat Politècnica de Catalunya (UPC), Barcelona 08034, Spain.}
}

\begin{document}

\maketitle 
\pagestyle{empty}  % no page number for the second and the later pages
\thispagestyle{empty} % no page number for the first page

\begin{abstract}
Endovascular interventions are high-stakes procedures requiring precise device operation within complex and tortuous vascular anatomies. Autonomous endovascular navigation has the potential to standardize procedural quality and reduce the performance variability inherent in manual operation.
Although Reinforcement Learning (RL) approaches have demonstrated promise in enabling autonomy in endovascular intervention, they often struggle with explicit constraint satisfaction and safety guarantees.
To address these challenges, a learning-based expert strategy is introduced, enhancing procedural consistency in autonomous endovascular intervention by explicitly decoupling high-level strategic decision-making from low-level procedural execution. The proposed framework replicates the expert clinical decision-making process: a strategic RL policy generates global navigation intents, which are subsequently refined through an expert-informed execution module. This module ensures that robot movements strictly adhere to expert operational norms, real-time kinematic limits, and vessel safety constraints. Experimental evaluation across high-fidelity 3D simulations and a real-world robotic platform demonstrates that the proposed framework not only outperforms baseline policies but also effectively replicates expert-level proficiency. The framework achieves a high navigation success rate ($> 96\%$) and a 29.3\% reduction in operational steps, which translates to enhanced operative efficiency and minimized device-vessel interaction. Furthermore, a 13\% reduction in trajectory variance indicates superior procedural standardization, aligning autonomous behavior with established clinical norms. These results underscore its potential to enhance the predictability, safety, and consistency of robotic endovascular interventions.
\end{abstract}

\section{Introduction}
Endovascular interventions are essential minimally invasive procedures for treating cardiovascular and neurovascular diseases \cite{liang2026self}. These procedures require the precise manipulation of flexible guidewires and catheters through tortuous vascular structures under fluoroscopic guidance \cite{pore2023autonomous, yao2025realrecon}. Meanwhile, the operator must interpret 2D fluoroscopy images in real time, anticipate guidewire deformation, and apply precise control to avoid vessel-wall contact \cite{yao2023enhancing}. Such demands impose a steep learning curve, prolonged procedure times, and sustained radiation exposure \cite{yao2025advancing}. Moreover, the variability in operator skill levels introduces significant inconsistency in procedural outcomes, underscoring the clinical need for intelligent robotic systems that can standardize interventional performance \cite{dupont2025grand}.

\begin{figure}[t]
\centering
\includegraphics[width=0.46\textwidth]{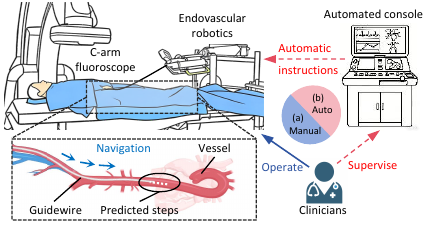}
\caption{The robot-assisted endovascular interventions are the minimally invasive procedures that utilize interventional guidewires, a C-arm X-ray machine, and endovascular robotics. (a) Manual operation: clinicians directly operate the robotic system based on their clinical expertise and procedural experience. (b) Autonomous operation: an automated console generates automatic instructions based on real-time fluoroscopes to endovascular robotics, where clinicians supervise the procedure. The lower panel illustrates guidewire navigation within complex vessel geometries, where predicted motion steps are executed toward the target region.} \vspace{-0.6cm}
\label{fig:background}
\end{figure}

Robot-assisted autonomous endovascular navigation has emerged as a promising approach to reduce operator workload and standardize procedural quality. Within the evolving landscape of intervention autonomy \cite{yao2025advancing, dupont2025grand}, reinforcement learning (RL) has emerged as a particularly promising paradigm, demonstrating the capacity to acquire guidewire navigation policies directly from simulated vascular environments~\cite{yao2025multi, yao2025sim4endor}, adapting to complex anatomical geometries and nonlinear device dynamics through high-dimensional visual observations~\cite{scarponi2024autonomous, yao2025realtracking,zhao2026vision,yao2025realrecon}. However, existing RL-based methods exhibit limitations in clinical viability. Pure learning-based policies are prone to locally optimal decisions that manifest as redundant maneuvers and excessive navigation steps \cite{wang2025learning}. More critically, the absence of explicit constraint enforcement renders these methods unable to guarantee compliant device behavior under the uncertain contact conditions characteristic of real endovascular procedures, leading to unpredictable and potentially risky outcomes.

Structured regulation through optimization-based frameworks provides a robust mechanism to ensure technical compliance and motion consistency. Model Predictive Control (MPC), in particular, enables physically feasible trajectory generation by embedding operational constraints within an online optimization process~\cite{li2023nonlinear}. Nevertheless, purely model-based formulations often struggle to capture the anatomy-dependent and highly nonlinear behavior of flexible guidewires, especially under conditions of incomplete sensing or unmodeled dynamics. Recent developments in autonomous systems have sought to unify these paradigms by coupling learning-based decision-making with optimization-based constraint regulation~\cite{romero2025actor}. These hybrid architectures have demonstrated significant gains in robustness in complex domains such as agile flight and autonomous driving, motivating the adaptation of similar integration strategies to endovascular interventions.

Building upon this foundation, a learning expert strategy is proposed for autonomous robotic endovascular intervention via a decoupled procedural execution framework. The design leverages an explicit separation of strategic navigation intent from execution refined by clinical expertise, ensuring that robot movements consistently satisfy safety constraints and expert operational skills. An RL policy captures global navigation strategies from visual observations, while a downstream predictive refinement module enforces kinematic and vessel boundary constraints through real-time optimization. This decoupling frees the learning process from simultaneously resolving long-horizon planning and hard constraint compliance, enabling more effective strategy discovery. At deployment, the refinement module filters policy outputs to produce smooth, constraint-compliant actions, mirroring interventionalists who delineate procedural intent from precise device manipulation.

The proposed framework is validated in a high-fidelity 3D vascular simulation environment and further deployed on a physical robotic endovascular platform. Experimental results demonstrate that the framework achieves reliable autonomous navigation with improved efficiency, reducing navigation steps and trajectory variance while maintaining a high success rate compared with baseline RL methods. These results demonstrate the effectiveness of decoupling learning-based strategy acquisition from model-based constraint enforcement, and demonstrate the potential of this paradigm for reliable autonomous endovascular intervention.

The main contributions of this work are as follows:

\begin{itemize}
\item A constraint-decoupled RL framework for autonomous endovascular guidewire navigation that separates strategic intent acquisition from operational constraint enforcement, providing a clinically aligned formulation for autonomous robotic endovascular intervention.

\item An online predictive motion refinement module that explicitly incorporates guidewire kinematics, curvature constraints, and vessel centerline adherence to ensure real-time constraint compliance and trajectory smoothness.

\item Comprehensive validation in both simulation and real-world robotic experiments, demonstrating improved navigation efficiency and trajectory consistency.
\end{itemize}

\begin{figure*}[!htbp]
\centering
\includegraphics[width=0.92\textwidth]{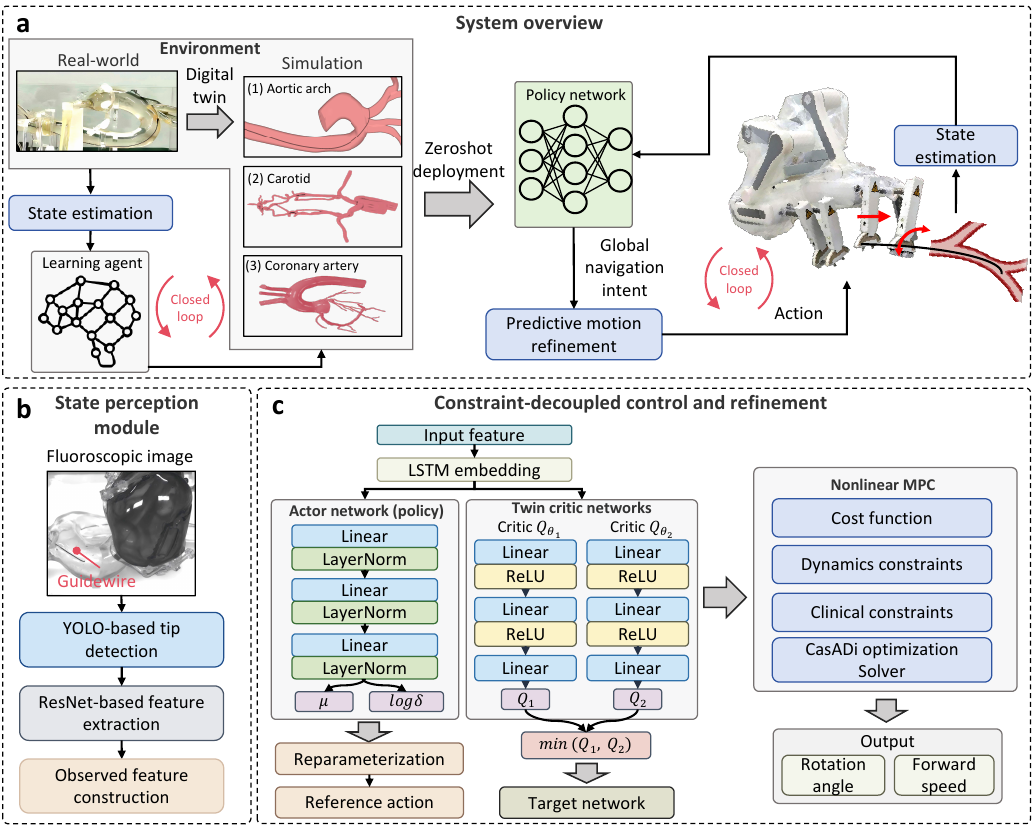}
\caption{Framework overview and working principle of the proposed constraint-decoupled RL-MPC framework for autonomous endovascular navigation. (a) An integrated architecture bridging digital twin-based policy optimization with real-world robotic deployment. (b) A vision-based module for real-time guidewire tip tracking and spatio-temporal feature representation. (c) The core constraint-decoupled scheme, where a high-level RL policy generates strategic intents, which are subsequently refined by a nonlinear MPC module to enforce kinematic and safety constraints.} \vspace{-0.6cm}
\label{fig:OverallFramework}
\end{figure*}

\section{Methodology}
\subsection{Problem Formulation}
The autonomous endovascular navigation task is formulated as a sequential decision-making problem under boundary constraints within a continuous vascular domain $\mathcal{V} \subset \mathbb{R}^3$. The vascular environment is characterized by the vessel boundary $\partial\mathcal{V}$ and the anatomical centerline $\mathcal{C} = \{\mathbf{p}_i\}_{i=1}^{N}$. Since clinical procedures are typically performed under 2D fluoroscopic guidance, the system state $\mathbf{x}_t \in \mathcal{X}$ at time $t$ is defined by the observable projected guidewire tip configuration:
\begin{equation}
    \mathbf{x}_t = [x_t, y_t, v_{x,t}, v_{y,t}, \theta_t]^T,
\end{equation}
where $(x_t, y_t)$, $(v_{x,t}, v_{y,t})$, and $\theta_t$ denote the tip position, velocity, and orientation in the imaging plane, respectively. The control input $\mathbf{u}_t \in \mathcal{U}$ represents the actuation applied at the proximal end of the guidewire by the robotic manipulator:
\begin{equation}
    \mathbf{u}_t = [v_{\text{insert},t}, \omega_t]^T,
\end{equation}
where $v_{\text{insert},t}$ is the insertion velocity and $\omega_t$ is the rotational velocity.

The primary objective is to generate an optimal control strategy that drives the guidewire from an initial state $\mathbf{x}_0$ to a target anatomical location $\mathbf{x}_{\text{goal}} \in \mathcal{C}$. This navigation process must minimize cumulative procedural costs while adhering to a set of operational constraints $\mathcal{S}_{\text{bound}}$, which include maintaining vessel wall clearance, adhering to curvature limits $\kappa_{\max}$ to prevent device deformation, and tracking the vascular centerline. This task is modeled as a Constrained Markov Decision Process (CMDP) and is addressed through a constraint-decoupled framework that separates global strategic intent from local motion refinement.
\subsection{Integrated Strategy-Execution Architecture}
The proposed system utilizes a dual-stage architecture designed to mirror the clinical decision-making paradigm of interventionalists. Rather than employing an end-to-end mapping, the framework separates the acquisition of strategic navigation intent from the enforcement of operational constraints. As illustrated in Fig. \ref{fig:OverallFramework}, the framework comprises three functional modules: a visual perception pipeline, a strategic navigation policy, and a predictive motion refinement module.

The process begins with the perception module, which extracts the observation vector $\mathbf{o}_t$ representing the guidewire state and environmental features from fluoroscopic imagery. These observations are provided to the strategic navigation policy $\pi_\phi$, which is trained in a high-fidelity 3D vascular simulation to learn global navigation tactics. The policy generates a reference control action, or navigation intent $\mathbf{u}_{\text{RL}} \sim \pi_\phi(\cdot | \mathbf{o}_t)$, aimed at maximizing long-term task success and procedural efficiency. To ensure rigorous adherence to technical constraints during execution, $\mathbf{u}_{\text{RL}}$ serves as the reference input to the downstream predictive refinement module. This module solves a finite-horizon optimization problem to filter the control actions, utilizing a simplified kinematic model $f(\mathbf{x}_k, \mathbf{u}_k)$ to predict future trajectories while enforcing explicit constraints on boundary clearance and curvature. The resulting optimal action $\mathbf{u}_0^*$ is subsequently executed by the robotic manipulator. This decoupling leverages the strategic adaptability for complex path planning while maintaining the operational consistency and smoothness provided by optimization-based control.
\subsection{Visual Perception and State Estimation}
Real-time state estimation is achieved through a vision processing pipeline that interprets fluoroscopic feedback. The system acquires images via industrial cameras to track the interactions between the guidewire, catheter, and sheath. To localize the guidewire tip position $\mathbf{p}_{\text{tip}} = (x_{\text{tip}}, y_{\text{tip}})$, a YOLOv5 object detection model is employed to provide bounding box coordinates, from which the pixel-level center is derived. To mitigate detection failures, a temporal consistency check utilizes the last known valid position when detection confidence falls below a set threshold.

High-dimensional visual features are extracted to provide context regarding the vascular structure. A ResNet50 convolutional neural network, with the final classification layer removed, processes the standardized input images to yield a 2048-dimensional feature vector $\mathbf{f}_{\text{img}}$ that implicitly encodes spatial relationships and anatomical textures. The composite observation vector $\mathbf{o}_t$ aggregates the explicit geometric state and implicit visual features:
\begin{equation}
    \mathbf{o}_t = [\mathbf{p}_{\text{tip}}, \mathbf{f}_{\text{img}}, \mathbf{u}_{t-1}, \mathbf{p}_{\text{target}}],
\end{equation}
where $\mathbf{u}_{t-1}$ is the previous control action and $\mathbf{p}_{\text{target}}$ is the goal position. This multi-modal representation enables the policy to reason about both the immediate tool configuration and the broader anatomical context.

\subsection{Strategic Navigation Policy}
The strategic navigation behavior is governed by a policy trained using Soft Actor-Critic (SAC), an off-policy algorithm that maximizes an entropy-regularized objective. The policy network $\pi_\phi(\mathbf{u}_t | \mathbf{o}_t)$ outputs a distribution over continuous navigation intents, while the Q-function networks $Q_\theta(\mathbf{o}_t, \mathbf{u}_t)$ estimate the expected return.

The training process is conducted within a simulated 3D vascular environment modeling complex aortic arch anatomy. The reward function $R(\mathbf{x}_t, \mathbf{u}_t)$ is designed to balance procedural completion with efficiency and boundary awareness. The total reward at each timestep is defined as:
\begin{equation}
    R = r_{\text{target}} + r_{\text{path}} + r_{\text{step}} + r_{\text{boundary}}.
\end{equation}
The target reaching term $r_{\text{target}}$ assigns a sparse positive reward $R_{\text{target}}$ upon arrival at the goal. Path efficiency is encouraged via $r_{\text{path}} = w_{\text{path}} \Delta d_{\text{path}}$, which rewards the reduction in geodesic distance along the centerline. A constant step penalty $r_{\text{step}}$ is applied to incentivize rapid completion. The boundary penalty $r_{\text{boundary}}$ imposes a negative cost whenever the guidewire tip violates the minimum safe distance $d_{\min}$ from the vessel wall. This penalty serves as a behavioral bias during training, guiding the policy toward the centerline while the downstream refinement module ensures strict constraint compliance during deployment.

\subsection{Predictive Motion Refinement}
To ensure technical compliance and kinematic feasibility, the reference intents generated by the policy, denoted as $\mathbf{u}_{\text{RL}}$, are then refined by a Nonlinear MPC layer. The module solves a constrained optimization problem over a finite prediction horizon $H$.

\subsubsection{Dynamics Model}
The prediction module employs a discrete-time formulation of guidewire kinematics. The state evolution $f(\mathbf{x}_k, \mathbf{u}_k)$ is defined as:
\begin{align}
    x_{k+1} &= x_k + v_{x,k} \Delta t, \\
    y_{k+1} &= y_k + v_{y,k} \Delta t, \\
    v_{x,k+1} &= \alpha v_{\text{insert},k} \cos(\theta_k) + (1-\alpha) v_{x,k}, \\
    v_{y,k+1} &= \alpha v_{\text{insert},k} \sin(\theta_k) + (1-\alpha) v_{y,k}, \\
    \theta_{k+1} &= \theta_k + \omega_k \Delta t,
\end{align}
where $\Delta t$ is the sampling time and $\alpha \in (0, 1)$ is a smoothing factor acting as a low-pass filter to model the inertial response of the physical device.

\subsubsection{Optimization Formulation}
At each control step $t$, the NMPC solves for the optimal sequence of control inputs $\mathbf{u}_{0:H-1}^*$ that minimizes the deviation from the strategic intent and the target trajectory. The optimization problem is formulated as:
\begin{equation}
\label{eq:mpc_objective}
\min_{\mathbf{u}_{0:H-1}} \sum_{k=0}^{H-1} (\|\mathbf{x}_k - \mathbf{x}_{\text{ref},k}\|^2_{\mathbf{Q}_{\text{ref}}} + \|\mathbf{u}_k - \mathbf{u}_{\text{RL}}\|^2_{\mathbf{R}}) + J_{\text{terminal}}(\mathbf{x}_H),
\end{equation}
where $\mathbf{x}_{\text{ref},k}$ is a target-directed reference trajectory, and $\mathbf{Q}_{\text{ref}}$ and $\mathbf{R}$ are weighting matrices. The term $\|\mathbf{u}_k - \mathbf{u}_{\text{RL}}\|^2_{\mathbf{R}}$ penalizes deviations from the learned strategic intent, ensuring the refinement module preserves the navigation tactics unless operational constraints require intervention. The optimization is subject to the following constraints for $k = 0, \ldots, H-1$:
\begin{align}
    \mathbf{x}_{k+1} &= f(\mathbf{x}_k, \mathbf{u}_k), \\
    \mathbf{u}_{\min} &\leq \mathbf{u}_k \leq \mathbf{u}_{\max}, \\
    \frac{|\omega_k|}{|v_{\text{insert},k}| + \epsilon} &\leq \kappa_{\max}, \\
    d_{\mathcal{C}}(\mathbf{x}_k) &\leq d_{\max},
\end{align}
where $\mathbf{u}_{\min}$ and $\mathbf{u}_{\max}$ are actuator limits, and $\kappa_{\max}$ represents the curvature threshold. The constraint $d_{\mathcal{C}}(\mathbf{x}_k) \leq d_{\max}$ enforces vessel boundary compliance. The optimization is implemented using the CasADi framework~\cite{Andersson2019} with the IPOPT solver~\cite{chiang2016inertia}, applying a receding horizon strategy that executes only the first action $\mathbf{u}_0^*$. This formulation effectively decouples strategic planning from real-time constraint enforcement.

\section{Experiments and Results}
This section presents the evaluation of the proposed MPC-constrained RL framework for autonomous endovascular guidewire navigation in both simulation and physical settings. Systematic navigation tasks and ablation studies performed in a 3D vascular simulation environment validate the performance of this framework for endovascular procedures. Then, the trained policy was deployed on a robotic endovascular platform for real-world assessments.
    
\subsection{Experimental Configuration} 
The simulation platform is built upon the Sim4EndoR environment~\cite{yao2025sim4endor}, which integrates SOFA v23.06~\cite{yao2025sim2real} with SofaPython3, BeamAdapter, and Blender. This modular setup supports the incorporation of patient-specific 3D vascular models and diverse robotic intervention scenarios. The guidewire is modeled using a beam-based finite element approach~\cite{BeamAdapter}, featuring a J-shaped configuration with a straight body segment (0.89\,mm diameter) and a reduced-diameter tip (0.7\,mm diameter). The mechanical stiffness is specifically tuned to match commercially available guidewires to ensure realistic navigation dynamics within tortuous vasculature.

Training and simulation evaluations are executed on a workstation equipped with an Intel® Core™ i9-14900HX CPU (16 cores) and an NVIDIA GeForce RTX 4060 GPU. The software implementation utilizes PyTorch 1.11.0 and Python 3.8. The strategic navigation policy is developed using the Gymnasium library~\cite{gymnasium}, while the predictive motion refinement module is implemented via the CasADi optimization framework.

\subsection{Strategy-Based Learning Configuration}

\subsubsection{Algorithm and Hyperparameters}
The strategic navigation policy is acquired through the Soft Actor-Critic (SAC) algorithm~\cite{haarnoja2018soft}, selected for its off-policy efficiency and entropy-regularized exploration, which is critical for handling the highly nonlinear dynamics of endovascular navigation. Key hyperparameters, selected to ensure stable convergence, are detailed in Table~\ref{tab:hyperparameters}.

\begin{table}[!htbp]
\centering
\caption{Hyperparameter Settings for Constraint-Decoupled Learning}
\label{tab:hyperparameters}
\renewcommand{\arraystretch}{1.2} % 稍微加宽行间距
\footnotesize
% 定义灰色条带，从第二行开始，奇数行白色，偶数行浅灰色
\rowcolors{2}{gray!7}{white} 
\begin{tabular}{p{4.5cm}l} % 固定左侧宽度，使表格比例更协调
\specialrule{1.2pt}{0pt}{0pt} % 加粗的顶线
\rowcolor{white} \textbf{Parameter} & \textbf{Value} \\
\midrule
Learning rate ($lr$) & $3.22 \times 10^{-4}$ \\
Discount factor ($\gamma$) & $0.99$ \\
Target update rate ($\tau$) & $0.005$ \\
Batch size & $32$ \\
Replay buffer size & $10^4$ episodes \\
Reward scaling & $1.0$ \\
LR end factor & $0.15$ \\
LR decay steps & $6 \times 10^6$ \\
LSTM hidden nodes & $700$ \\
MLP architecture & $3 \times 400$ \\
Entropy temperature ($\alpha$) & Auto-adjusted \\
\specialrule{1.2pt}{0pt}{0pt} % 加粗的底线
\end{tabular}
\end{table}

\subsubsection{Network Architecture}
The policy and twin Q-function networks are implemented as multi-layer perceptrons (MLPs) integrated with a shared Long Short-Term Memory (LSTM) embedder. This architecture captures temporal dependencies in the observation stream $\mathbf{o}_t$. The shared LSTM processes raw observations into an embedded state $\mathbf{z}_t \in \mathbb{R}^{700}$, which is then passed to the policy network comprising three hidden layers of 400 neurons each with ReLU activations. The output layer parameterizes a Gaussian distribution via $\boldsymbol{\mu}_\phi(\mathbf{z}_t)$ and $\log \boldsymbol{\sigma}_\phi(\mathbf{z}_t)$. The twin Q-networks mirror this MLP structure, utilizing clipped double Q-learning to mitigate overestimation bias.

\subsubsection{Reward Hierarchy and Decoupling}
The reward function is structured to balance procedural intent and navigation efficiency. The weighting terms follow the hierarchy $w_{\text{target}} \gg w_{\text{path}} \gg |w_{\text{step}}|$, ensuring that task completion remains the primary learning signal while path efficiency and step minimization act as secondary shaping terms. Notably, in accordance with the constraint-decoupled paradigm, vessel-wall boundary constraints are not explicitly enforced within the reward function during the phase. This design allows the strategic policy to focus exclusively on acquiring high-level navigation tactics, delegating boundary compliance and technical regulation to the downstream refinement module.

\subsection{Simulation Evaluation}
Quantitative evaluation was conducted in an endovascular intervention simulator to assess the efficacy of the proposed navigation paradigm. All experiments were performed over 50 navigation episodes with identical initialization conditions to ensure statistical consistency.

\subsubsection{Ablation on Refinement Weighting Factor}
An ablation study was conducted by varying the normalized refinement weight $\lambda_{\text{MPC}}$ from $0.0$ to $1.0$ to identify the optimal balance between policy intent and constraint refinement.
Across the interval $\lambda_{\text{MPC}} \in [0.0, 0.5]$, the navigation success rate consistently remained above 85\%, indicating that moderate refinement authority does not compromise task completion. However, when $\lambda_{\text{MPC}}$ exceeded 0.5, performance deteriorated markedly, with the success rate decreasing from 70\% to 40\% as $\lambda_{\text{MPC}}$ increased further. 
This degradation suggests that excessive dominance of the optimization layer suppresses the learned strategic policy. In this regime, the control signal is primarily governed by local constraint satisfaction, effectively reducing the system to a tracking-oriented controller lacking sufficient global navigation guidance. This behavior aligns with the architectural principle of the proposed framework, where the RL policy generates high-level intent and the MPC module functions as a subordinate feasibility refinement layer rather than a primary decision-maker.
Fig.~\ref{fig:weightcompare}(a) reports results within the representative range $[0, 0.5]$, where navigation performance exhibits a trade-off between policy autonomy and constraint authority. When $\lambda_{\text{MPC}}$ is small ($<0.2$), the optimization layer provides limited corrective influence, resulting in behavior similar to the baseline RL policy, including inefficient oscillatory adjustments near bifurcations. Conversely, when $\lambda_{\text{MPC}}$ is large ($>0.4$), the MPC layer dominates the control signal, which tends to suppress the learned navigation strategy and reduce adaptability to complex vessel geometries. An intermediate value of $\lambda_{\text{MPC}}=0.1$ achieved the highest success rate and optimal navigation efficiency; this value was therefore selected for all subsequent comparisons.

\subsubsection{Trajectory Consistency Analysis}
\begin{figure*}[!htbp]
\centering
\includegraphics[width=0.96\textwidth]{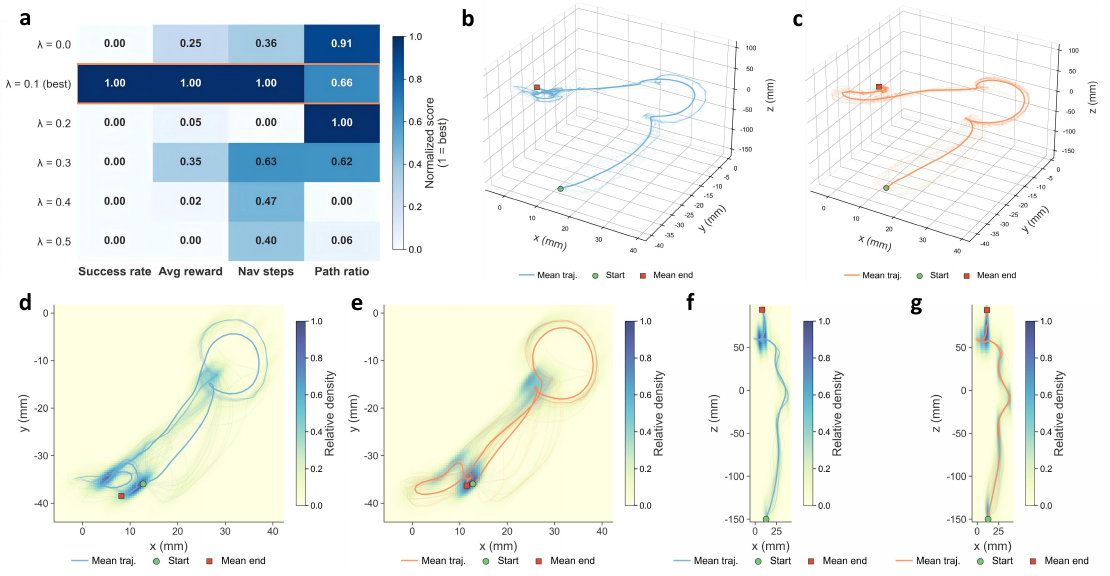}
\caption{Experimental results of autonomous guidewire navigation in simulated vascular environments. (a) Heatmap of normalized performance scores (1 indicates best) across four metrics: success rate, average episodic reward, navigation steps, and path ratio; $\lambda_{\mathrm{MPC}}=0.1$ yields the most balanced overall performance. (b,c) Representative 3D guidewire tip trajectories for (b) the baseline RL policy and (c) the proposed MPC constrained RL framework; solid curve denotes the mean trajectory over repeated rollouts ($N=20$), with the start position shown as a green circle and the mean terminal position shown as a red square. (d,e) Spatial visitation density of guidewire tip positions in the $x$--$y$ imaging plane accumulated over all time steps and rollouts ($N=20$) for (d) baseline RL and (e) MPC constrained RL; the overlaid solid curve is the mean trajectory. (f,g) Corresponding $x$--$z$ projections of the visitation density for (f) baseline RL and (g) MPC constrained RL. The goal region is a fixed sphere of 5\,mm radius; episodes terminate upon entry, so final positions vary within the sphere, resulting in small differences in mean terminal location.}
\label{fig:weightcompare}
\end{figure*}

Trajectory density was evaluated under stochastic rollouts from identical start and goal configurations to quantify the effect of the optimization layer on procedural repeatability. Superimposed 3D trajectories and probability density heatmaps in Fig.~\ref{fig:weightcompare}(b-g) illustrate that the integrated framework produces a noticeably more concentrated distribution compared to the diffuse patterns of the baseline. Quantitative metrics in Table~\ref{tab:trajectory_density} confirm this observation, with the lateral trajectory width in the $x$--$y$ plane ($\sigma_{\text{w,xy}}$) decreasing by $11.4\%$. Additionally, the spatial spread along the primary navigation axis ($\sigma_x$) was reduced by $13.0\%$, indicating that the framework yields more consistent and predictable trajectories.

\begin{table}[!htbp]
\centering
\caption{Comparison of trajectory consistency metrics.}
\label{tab:trajectory_density}
\begin{tabular}{lcccccc}
\toprule
\textbf{Method} & $\sigma_x$ & $\sigma_y$ & $\sigma_z$ & $\sigma_{\text{w,xy}}$ & $\sigma_{\text{w,xz}}$ & $\sigma_{\text{w,3D}}$ \\
\midrule
Baseline RL & $16.86$ & $11.77$ & $82.50$ & $6.50$ & $13.98$ & $17.59$ \\
Ours & $14.67$ & $11.02$ & $78.87$ & $5.76$ & $13.11$ & $17.11$ \\
\midrule
Reduction & $13.0\%$ & --- & $4.4\%$ & $11.4\%$ & $6.2\%$ & $2.7\%$ \\
\bottomrule
\end{tabular}
\vspace{2pt}
\begin{flushleft}
\footnotesize \textit{Note: All $\sigma$ values are in mm. Lateral spreads $\sigma_{w,xy}$ and $\sigma_{w,xz}$ are derived via PCA (lower is better).}
\end{flushleft}
\end{table}

\begin{table*}[t]
\centering
\caption{Comparison of Performance Evaluations in Endovascular Navigation Tasks Across Multiple Anatomical Geometries.}
\label{tab:simulation_results}
\resizebox{\textwidth}{!}{
\begin{tabular}{l|ccc|ccc|ccc}
\toprule

\multirow{2}{*}{Metric} &
\multicolumn{3}{c|}{Brachiocephalic} &
\multicolumn{3}{c|}{Carotid} &
\multicolumn{3}{c}{Subclavian} \\

\cline{2-10}

& SAC & DDPG & Ours &
  SAC & DDPG & Ours &
  SAC & DDPG & Ours \\

  \midrule

  \rowcolor{gray!15}
  Success Rate $\uparrow$ & 92\% (46/50) & 86\% (43/50) & 98\% (49/50) & 94\% (47/50) & 92\% (46/50) & 96\% (48/50) & 90\% (45/50) & 84\% (42/50) & 98\% (49/50) \\
  
  Steps $\downarrow$ & 85.89 $\pm$ 18.23 & 88.51 $\pm$ 9.89 & 71.55 $\pm$ 10.14 & 94.24 $\pm$ 14.65 & 118.24 $\pm$ 25.62 & 71.74 $\pm$ 3.62 & 98.75 $\pm$ 8.21 & 128.74 $\pm$ 13.21 & 70.92 $\pm$ 5.18 \\
  
  \rowcolor{gray!15}
  Path Ratio $\uparrow$ & 0.957 $\pm$ 0.075 & 0.978 $\pm$ 0.032 & 0.984 $\pm$ 0.008 & 0.965 $\pm$ 0.069 & 0.959 $\pm$ 0.019 & 0.981 $\pm$ 0.016 & 0.982 $\pm$ 0.007 & 0.985 $\pm$ 0.023 & 0.986 $\pm$ 0.005 \\
  
  Reward $\uparrow$ & 2.198 $\pm$ 0.195 & 2.205 $\pm$ 0.186 & 2.215 $\pm$ 0.056 & 2.123 $\pm$ 0.165 & 2.105 $\pm$ 0.086 & 2.142 $\pm$ 0.074 & 2.038 $\pm$ 0.112 & 2.045 $\pm$ 0.074 & 2.221 $\pm$ 0.064 \\
  
  \bottomrule
  \end{tabular}}
\end{table*}

\subsubsection{Comparative Study on Navigation Performance Across Multiple Anatomical Geometries.}
Using the selected refinement weight, the proposed framework was compared against pure RL policies trained under identical settings. As summarized in Table~\ref{tab:simulation_results}, the integrated framework consistently improves both navigation safety and efficiency across the Brachiocephalic, Carotid, and Subclavian anatomies.
Aggregated across three geometries, the success rate is increased to above 96\%, demonstrating enhanced robustness and stability. Meanwhile, the average procedural steps are reduced by 29.3\% compared to baseline RL policies, with a statistically significant difference confirmed by Welch’s t-test (both $p < 0.003$), indicating improved procedural efficiency. Furthermore, the higher path ratio and reward suggest closer adherence to the vessel centerline, reflecting safer and smoother trajectories.
These results indicate that the proposed strategy not only increases task success rate but also improves procedural efficiency while maintaining high reliability.

\begin{figure*}[!htbp]
\centering
\includegraphics[width=0.90\textwidth]{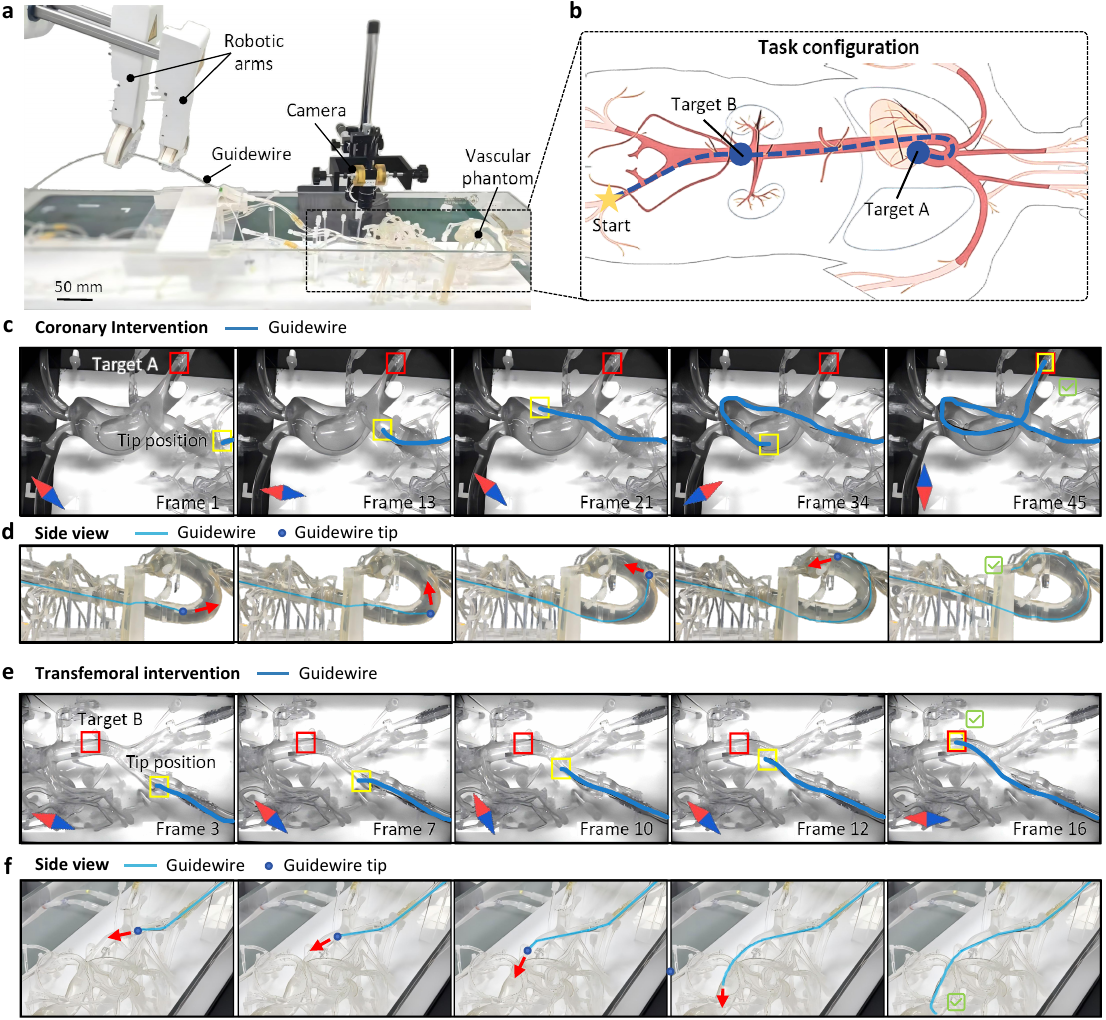}
\caption{In vitro phantom experiments for autonomous guidewire navigation. (a) Bench-top experimental setup, where a robotic actuation module advances and manipulates a clinical guidewire inside a patient-specific vascular phantom under overhead vision (scale bar, 50\,mm). (b) Task configuration with a fixed entry/start location and two target sites (Target~A and Target~B). Starting from the proximal access, the guidewire is inserted into the phantom lumen, navigates through bifurcations, and is guided to the selected distal target. (c,d) Representative time-lapse frames of \textit{coronary intervention} navigation. From the start, the guidewire tip is progressively advanced along the main lumen, and finally reaches the target region (Target~A). (c) Imaging-plane view and (d) synchronized side view. The tracked guidewire centerline is overlaid in blue; the red box denotes the target region and the yellow box indicates the current tip position. A green checkmark marks successful target reaching. (e,f) Representative time-lapse frames of \textit{transfemoral intervention} navigation. The guidewire is introduced from the proximal access, traverses the vessel tree, and is steered into the branch leading to Target~B. (e) Imaging-plane view and (f) synchronized side view, using the same visual conventions as in (c,d).}
\label{fig:deployment}
\end{figure*}

\subsection{In vitro Validation on Vascular Phantoms}
The proposed framework was deployed on the ALLVAS\texttrademark{} robotic system (Shanghai Operation Robot Co., Ltd., Shanghai, China), which integrates a precision manipulator with a dedicated guidewire manipulation mechanism. The vision system utilized a high-resolution camera (Hikvision, Hangzhou, China) for real-time feedback, employing YOLOv5 for the detection of the guidewire tip, catheter, and sheath, while ResNet50 was used for visual feature extraction. In this deployment, the RL policy generated high-level navigation intent based on visual observations, which was then refined by a CasADi-based MPC controller. The MPC layer enforced constraints including vessel-boundary clearance thresholds, curvature limits, and velocity bounds tailored to the physical properties of the guidewire.

Validation was conducted through 20 navigation trials within a silicone vascular phantom maintained at 36.8\,$^{\circ}$C by a circulation system to simulate physiological conditions (Fig.~\ref{fig:deployment}). The framework achieved a $90.0\%$ success rate (18 out of 20 trials), with an average navigation time of $4.2 \pm 0.8$ minutes and $156 \pm 32$ control steps per trial. Although a moderate performance drop was observed compared to simulation results, this is consistent with the expected sim-to-real gap caused by perception uncertainty. Notably, zero vessel-boundary violations were recorded across all physical experiments, as the MPC layer successfully intervened during potential buckling or near-boundary events.

\section{Discussion}
This study proposes an integrated navigation framework that combines learning-driven strategy acquisition with predictive, optimization-based control to address the trade-off of efficiency and precision in autonomous endovascular interventions, specifically the requirement for precise constraint satisfaction within confined and tortuous vascular anatomies. While conventional learning-based methods offer high strategic adaptability, they often struggle to maintain strict boundary adherence and exhibit unstable oscillatory behaviors near sharp bifurcations. Experimental results demonstrate that the proposed optimization-based refinement layer effectively resolves these limitations by filtering high-level policy commands through kinematic and boundary constraints. This synergy ensures superior trajectory consistency and motion smoothness, providing a robust mechanism to maintain predictable device behavior within delicate anatomical spaces.

The architecture mirrors the multi-stage decision logic employed by interventionalists who separate global procedural planning from the nuanced manipulation of instruments. By replicating this workflow, the proposed strategy provides an expert-informed mechanism for maintaining consistent vessel boundary clearance without compromising navigation goals. Furthermore, the modular design allows optimization constraints to be adjusted for diverse device configurations or patient-specific anatomies without necessitating policy retraining, establishing a versatile foundation for autonomous interventional robotics.

The transition from simulation to the physical platform resulted in a reduction in success rate from 98.0\% to 90.0\%, which may be attributable to unmodeled physical factors such as non-linear friction, mechanical hysteresis, and visual occlusions. Nevertheless, the persistent zero-violation rate in physical trials demonstrates that the optimization layer effectively mitigates policy-level deviations, maintaining procedural safety in real-world settings. Future efforts will prioritize the integration of high-fidelity biomechanical models, including vessel wall compliance and complex guidewire deformation, to further bridge the sim-to-real gap. Subsequent in vivo studies will evaluate the framework's clinical efficacy and robustness under dynamic conditions.

\section{Conclusion}
This study introduces a learning expert strategy designed to optimize the consistency and efficiency of autonomous endovascular navigation. By decoupling strategic decision-making from procedural execution, the system ensures strict adherence to expert-aligned procedural standards and vessel safety constraints while achieving superior operational smoothness. This architecture inherently replicates the expert decision-making process, where interventionalists distinguish high-level procedural intent from fine-grained device manipulation, thereby providing a robust mechanism to mitigate intraoperative risks. Such alignment with established interventional workflows underscores the potential of the proposed method for enhancing the precision and standardization of robotic interventions.

The modular structure of this framework enables its adaptation to various vascular geometries and device configurations, providing a scalable foundation for autonomous intervention robotics. To further refine this approach, future research will integrate advanced biomechanical models, including vessel wall compliance and non-linear guidewire deformation, to enhance the simulation-to-reality transition. Furthermore, in vivo animal trials need to be conducted to evaluate the clinical performance and translation potential within complex, dynamic physiological environments.

\bibliographystyle{ieeetr}
\balance
\bibliography{reference}
\end{document}